# Title page

1. **Title:** LOGICAL: Local Obfuscation by GLINER for Impartial Context-Aware Lineage – Development and evaluation of PII Removal system.
2. **Running title:** GLiNER for PII Removal
3. **Author names and affiliations:**

|   | Name | Academic qualifications | Department | Institution | City and country |
|---|------|-------------------------|------------|-------------|------------------|
| 1 | Prakrithi Shivaprakash | MD, PDF, DM | Department of Psychiatry | National Institute of Mental Health and Neuro Sciences (NIMHANS) | Bengaluru, India |
| 2 | Lekhansh Shukla* | MD, PDF, DM | Centre for Addiction Medicine, Department of Psychiatry | National Institute of Mental Health and Neuro Sciences (NIMHANS) | Bengaluru, India |
| 3 | Animesh Mukherjee | B-Tech, M-Tech, PhD (Computer Science and Engineering) | Department of Computer Science and Engineering | Indian Institute of Technology, Kharagpur (IIT-KGP) | Kharagpur, India |
| 4 | Prabhat Chand | MD, DNB, MNAMS | Centre for Addiction Medicine, Department of Psychiatry | National Institute of Mental Health and Neuro Sciences (NIMHANS) | Bengaluru, India |
| 5 | Pratima Murthy | DPM, MD, FRCP (Glasgow) | Centre for Addiction Medicine, Department of Psychiatry | National Institute of Mental Health and Neuro Sciences (NIMHANS) | Bengaluru, India |

4. **Corresponding author\***
   - Name: Lekhansh Shukla
   - Full postal address: Office of the Centre for Addiction Medicine, 2nd floor, Centre for Addiction Medicine Ward – Female wing, Department of Psychiatry, National Institute of Mental Health and Neuro Sciences (NIMHANS), Bengaluru – 560029, India.
   - Email address: drlekhansh@gmail.com
   - Telephone number: +91 9886160956.

**Conflicts of interest statement:**

- **Source of funding:** LS received an extramural grant from the Indian Council of Medical Research (ICMR), Government of India.
- **Conflicts of interest:** The authors have no conflicts of interests to declare.
- **Constraints on publishing:** None.

5. **Author contribution statement:**

Each author certifies that their contribution to this work meets the standards of the International Committee of Medical Journal Editors.

|   | Author | Contribution |
|---|--------|--------------|
| 1 | Prakrithi Shivaprakash | Conceptualisation, data collection, data analysis and manuscript preparation, interpretation of data, has approved the final manuscript. |
| 2 | Lekhansh Shukla | Conceptualisation, data collection, data analysis and manuscript preparation, interpretation of data, has approved the final manuscript. |
| 3 | Animesh Mukherjee | Conceptualisation, critical revision of manuscript, interpretation of data, has approved the final manuscript. |
| 4 | Prabhat Chand | Critical revision of manuscript, interpretation of data, has approved the final manuscript. |
| 5 | Pratima Murthy | Critical revision of manuscript, interpretation of data, has approved the final manuscript. |

# LOGICAL: Local Obfuscation by GLINER for Impartial Context-Aware Lineage – Development and evaluation of PII Removal system

## Abstract

**Background:** Removing Personally Identifiable Information (PII) from clinical notes in Electronic Health Records (EHRs) is essential for research and AI development. While Large Language Models (LLMs) are powerful, their high computational costs and the data privacy risks of API-based services limit their use, especially in low-resource settings. To address this, we developed LOGICAL (Local Obfuscation by GLINER for Impartial Context-Aware Lineage), an efficient, locally deployable PII removal system built on a fine-tuned Generalist and Lightweight Named Entity Recognition (GLiNER) model.

**Methods:** We used 1515 clinical documents from a psychiatric hospital's EHR system. We defined nine PII categories for removal. A `modern-gliner-bi-large-v1.0` model was fine-tuned on 2849 text instances and evaluated on a test set of 376 instances using character-level precision, recall, and F1-score. We compared its performance against Microsoft Azure NER, Microsoft Presidio, and zero-shot prompting with Gemini-Pro-2.5 and Llama-3.3-70B-Instruct.

**Results:** The fine-tuned GLiNER model achieved superior performance, with an overall micro-average F1-score of 0.980, significantly outperforming Gemini-Pro-2.5 (F1-score: 0.845). LOGICAL correctly sanitised 95% of documents completely, compared to 64% for the next-best solution. The model operated efficiently on a standard laptop without a dedicated GPU. However, a 2% entity-level false negative rate underscores the need for human-in-the-loop validation across all tested systems.

**Conclusion:** Fine-tuned, specialised transformer models like GLiNER offer an accurate, computationally efficient, and secure solution for PII removal from clinical notes. This "sanitisation at the source" approach is a practical alternative to resource-intensive LLMs, enabling the creation of de-identified datasets for research and AI development while preserving data privacy, particularly in resource-constrained environments.

## Background

Electronic Health Records (EHRs) serve as a comprehensive repository of medical information and are crucial for coordinating patient care within and across medical institutions. The digitisation of this clinical data also improves its research value by enabling the creation of large, shareable datasets. A significant part of these datasets comprises unstructured clinical notes, which provide detail and nuance that structured data cannot include, especially in fields like psychiatry (1). Advances in natural language processing (NLP) solutions have played a key role in unlocking the research potential of this unstructured text. Large language models (LLMs) are particularly effective in tasks such as summarising text, extracting structured information from unstructured notes, and are increasingly used to support clinical tasks such as clinical decision support systems (CDSS), mental healthcare chatbots, and analysing complex, large-scale data such as genomic and radiological information (2, 3).

Clinical notes may contain Personally Identifiable Information (PII) and Protected Health Information (PHI). Maintaining patient confidentiality and data privacy is a vital ethical and legal responsibility, regulated by laws such as the Health Insurance Portability and Accountability Act (HIPAA) in the United States, the General Data Protection Regulation (GDPR) in the European Union, and the Digital Personal Data Protection (DPDP) Act in India.

To support collaborative research, clinical data must undergo thorough de-identification before sharing and use. This process involves balancing the complete removal of PII with maintaining enough clinical context. For instance, while a specific date (including day, month, and year) acts as an identifier, a time range indicating symptom onset offers crucial clinical information and should typically be kept. However, traditionally, PII removal methods have been optimised for recall rather than precision. We need to consider downstream applications and find an appropriate balance between the two.

Clinical data is now also used to train large language models (LLMs) on various diagnostic and therapeutic tasks (Jiang et al., 2023). This calls for a re-evaluation of how we consider PII removal. Firstly, indiscriminate removal of information can compromise the data's value for training LLMs. Secondly, beyond confidentiality, this raises significant issues regarding biases and the reinforcement of historical inequalities (Suenghataiphorn et al., 2025). Consequently, information traditionally not considered PII—such as racial, linguistic, religious, or caste-related identities—also becomes important. We want models to learn clinical patterns without internalising biases. One solution is to replace PII in a detailed, lineage-preserving manner. This approach enables lineage-preservation, where each unique entity (e.g., a specific person's name) is consistently replaced by the same unique placeholder (e.g., 'Person_1') within and across all documents, allowing for meaningful downstream analysis. For example, if drug use is more common in a particular section of society, the related identifier could be replaced in clinical text with a neutral, unique, and consistent placeholder. This approach enables the model to learn associations without reinforcing pre-existing biases or becoming vulnerable to jailbreak attacks that aim to provoke LLMs into producing hateful or polarising content (Ackerman & Panickssery, 2025).

Automated methods are crucial for performing de-identification at scale, as manual redaction is both costly, time-consuming and prone to errors. Early automated solutions using rule-based systems, although effective for consistently formatted data such as phone numbers, fail to cope with variations like spelling or spacing mistakes. To address the limitations of rigid rule-based systems, this issue is framed as a Named Entity Recognition (NER) task, where machine learning models learn the contextual and linguistic patterns of identifiers from manually annotated notes (4).

The introduction of transformer architectures, such as Bidirectional Encoder Representations from Transformers (BERT), significantly improved performance on this task in clinical data (5). By employing self-attention mechanisms to process entire text sequences simultaneously, BERT models effectively interpret the unique jargon and abbreviations found in diverse clinical settings, while maintaining context. However, they require accurately annotated datasets for training and fine-tuning. More recently introduced LLMs like Generative Pre-trained Transformers (GPT) can perform complex tasks using zero-shot or few-shot prompting, potentially eliminating the need for large, task-specific annotated datasets for fine-tuning (6). Zero-shot refers to simply prompting an instruction-tuned LLM to extract all names of persons from a note. This is transformative because of its potential to extract fine-grained categories in a single pass.

Although LLMs appear to be an ideal solution, their deployment presents significant practical barriers, particularly in low-resource settings.

First, hosting a state-of-the-art LLM locally requires intensive computational resources. For instance, deploying the Llama-3.3-70B model (7), used in LLM-Anonymiser (6) will require high-end Graphics Processing Units (GPU). At standard 16-bit precision, the model alone consumes approximately 140 GB of Video Random Access Memory (VRAM) – 70 billion parameters * 2 bytes/parameter - and additional memory for effective processing of clinical notes, necessitating at least two NVIDIA A-100-80GB cards, which can cost upwards of $15,000 each. This is highlighted by *Wiest et al.* using quantised versions of the models, which can be utilised with limited computing resources. Nevertheless, LLM processing requires substantial time, energy, and computing resources.

Second, using an Application Programming Interface (API) to access the model, while convenient, introduces significant data security risks. This approach exposes sensitive patient data to third-party servers with potentially ambiguous data-use agreements, which may allow data containing PII or PHI to be used for training and may inadvertently be embedded in the model, potentially revealing it to others. In contrast, a locally deployed model not only ensures data security but also allows for fine-tuning on institution-specific data, thereby improving accuracy in local jargon, cultural, and demographic contexts.

Finally, most deep-learning solutions are inherently probabilistic in their output, meaning they can perform differently on the same task even under identical conditions. For high-stakes scenarios like PII masking, a human-in-the-loop (HIL) approach is therefore crucial. In such a setup, the most efficient design involves integrating a PII masking pipeline with EHR data entry. The free text is sanitised at source before being stored in a database, and the healthcare worker entering the data can act as the human-in-the-loop. Such a solution is only feasible if it can operate efficiently on edge devices, such as laptops.

These challenges highlight the critical need for a de-identification system that can operate securely "at the source", within an institution's own infrastructure, without demanding prohibitive computational resources, tailored to the identifier requirements of the particular institution.

We aver that we can use the benefits of progress in deep learning and transformer architecture without having to use LLMs for this task. For example, Generalist and Lightweight Named Entity Recognition (GLiNER) models (8). These models use modern-BERT (9) to encode the text (clinical text in this case) and another model to encode the desired entity labels (for example the word 'person'). This architecture allows zero-shot extraction of arbitrary entities based on semantic similarity between the entity label and the word to be classified (8). Furthermore, these are compact models that can be used on edge devices. However, to our knowledge there has been no research on utility of these models for PII-removal from real-world clinical notes.

In this paper, we demonstrate the efficacy of this approach by fine-tuning a GLiNER model on real-world, human-annotated clinical notes from a psychiatric institution.

This study demonstrates that fine-grained and time-efficient de-identification is achievable by fine-tuning transformer-based models with a relatively small dataset (1000 to 2000 documents). Our primary contribution is an openly available GLiNER model specifically optimised for fine-grained PII detection in real-world clinical notes. To facilitate its adoption, we also release an open-source library that allows institutions to apply this model directly to their text data to identify nine distinct entity types (person, address, address-state, address-country, dates, company, groups, languages and identification numbers) and replace them with placeholders which maintain context within and across notes.

# Methodology

## Data Source

The data for this study are sourced from the free-text entries in the EHR system of a tertiary-level teaching hospital. The entries include discharge summaries and consultation notes entered between January 2018 and December 2023. We used stratified sampling based on the number of words in each document. To arrive at a sample size of 1500 unique documents, 350 notes were sampled from each quantile. A smaller dataset of 15 transcripts of audio-recorded interviews of patients with substance use disorders (SUD) was also used. **The total number of unique, real-world documents is thus 1515.**

## Data Preprocessing

To reflect a real-world deployment scenario, we did not perform any standardisation or normalisation of the text. However, BERT-based models can only process up to 8000 tokens at a time. Therefore, we divided the original documents into sections of 4000 words or fewer (including punctuation), with a 25-word overlap on each side to preserve context. This step generated 3,108 instances, and for clarity, we will refer to these instances as the sample throughout the remainder of the manuscript.

## Definition and extraction of PII

In this study, we defined PII through a two-step empirical process. Two psychiatrists reviewed 100 randomly selected documents to identify all words or phrases meeting either of two criteria. First, words or phrases that can directly identify a person on their own or when combined with other information in the document. Second, if the words/phrases can bias a layperson reading the documents into believing that a particular psychiatric condition is associated with a specific group identity.

At the second step, we reviewed and discussed the words or phrases identified by the two psychiatrists. A key disagreement was about dates—whether to remove incomplete date mentions like "November 2020" or durations such as "in the last 10 days." Another disagreement concerned socioeconomic status, which is usually noted as middle or lower socioeconomic status. To reach a decision, we redacted all these phrases and read the note to see if it still conveyed clinical information and rationale behind the doctor's decision of locus-of-care or medications. We found that indiscriminately masking all date and time references made the note difficult to understand, as details like the sequence of events and age at onset are essential for diagnosis. Similarly, removing socioeconomic background obscured the reasoning behind medication choices and care location decisions. Since these two types of entities do not directly lead to identification, even when combined with information such as "this dataset originates from a particular hospital in a particular city," we decided not to include them. We recognise that these considerations may vary depending on the setting and team. We also recognise that we are not covering all the PII or PHI fields required for HIPAA or GDPR compliance.

Finally, the following 10 categories of PII are studied in the current work:

1. Name of persons (person).
2. Name of companies or organisations. This includes healthcare facilities as well as other companies, which can be employers, etc. (company).
3. Name of languages (language).
4. Fully specified dates or date ranges (dates).
5. Names of countries (address country).
6. Names of states or equivalent sub-national entities (address state).
7. All other locations or geographical entities that are smaller than a state (address).
8. Numeric or alphanumeric identifiers which can be directly linked to an individual, including phone numbers, voter's identification, driving license, etc. (identification number).
9. Names of groups that can bias or lead to identification, including religion, caste, tribes, political groups and self-help groups (groups).
10. Email or URL address. URL address to be marked only if it reveals the identity of a person (email URL).

Two nurses and one doctor annotated the dataset. We used Label Studio community edition to present the text to annotators. Two annotators independently completed all annotations; in cases of disagreement, a third annotator also annotated that instance to reach a majority vote. These instances with disagreements were then discussed during weekly review meetings.

Label studio allows 'limit selection to words', and this was used to prevent incomplete selection.

# Data Augmentation: Synthetic Data conditioned on real-world data

To enhance the generalisability of the model, we selectively augmented the data, focusing on the names of countries and states. To achieve this, we generated 117 synthetic instances by prompting the Llama-3.1-70B-instruct (10) model, conditioning on a randomly chosen existing instance (Supplementary Figure 01). The prompt instructed the model to expand or rephrase the clinical note to enable the insertion of at least one state name and one country name. These placeholders were replaced with a curated list of Indian states (including abbreviations and common misspellings) and countries.

These synthetic instances were also annotated in the same manner as real-world instances. With the addition of these 117 instances, **we now have a total of 3,225 instances**.

# Model Selection and Training

We compared two models based on GLiNER architecture – 'modern-gliner-bi-large-v1'(11) and 'NuNER_Zero-4k' (12) on 120 instances and chose the former based on its better performance.

The dataset was split into training (2849, 88%) and test (376, 12%) using random sampling.

We did a full finetuning of 'modern-gliner-bi-large-v1.0' (11). With only 0.53 billion trainable parameters, the model can be trained on a single consumer-grade GPU with 48 GB of RAM. The total training time for eight epochs was 65 minutes. Detailed training setup and the rationale behind the choice of hyperparameters are outlined in the Supplementary Table 01.

## Performance Evaluation and Comparisons

We have chosen four comparators – Microsoft Azure Text Analytics NER service (13), Microsoft Presidio (14), Gemini-Pro-2.5 (15) and Llama-3.3-70B-Instruct (7).

We chose Microsoft's NER service over PII because NER encompasses all PII fields and enables more fine-grained tagging of address and location-related entities. However, neither the NER nor the PII service has comparable outputs for three of our categories of interest – languages, identification numbers, and groups. Therefore, these three are excluded when comparing our model's performance with Microsoft Azure NER. Microsoft Presidio is an on-premise solution, but it lacks support for languages, and this category is excluded when comparing model performance. Supplementary tables 02 and 03 list the mapping between the Microsoft NER service, Microsoft Presidio and our labels.

We developed an optimised prompt for Gemini-Pro-2.5 (15) and Llama-3.3-70B-Instruct (7) to detect the nine types of entities of interest to us (Supplementary Table 04). The model's outputs were programmatically tagged in the input text using exact string matching.

To ensure parity while avoiding the transmission of PII to Azure and Gemini-Pro-2.5 API, two doctors pseudonymised the test set using a set of rules (Supplementary Table 05). They replaced seven types of PII under study (excluding language and groups) with appropriate pseudonyms.

For all evaluations and comparisons, we use character-level matching, yielding recall, precision, F1-score and Area Under the Receiver Operating Curve (AUROC). Micro and macro-averages are both reported. However, we use micro-averages as the primary statistic of comparison. Micro-average refers to the calculation of performance metrics from the aggregate sum of all true positives, false positives, and false negatives across the entire dataset. This method ensures that every individual prediction instance contributes equally to the final score, thereby reflecting the overall model performance.

We also report the False Negative Rate (FNR) at the entity level, meaning the proportion of entities that the models failed to detect entirely (missed all characters). This is important because even partial detection and flagging can catch a human's attention, whereas completely missing an entity might go unnoticed.

Finally, we report the number and proportion of instances where all entities were perfectly identified by various solutions (i.e. false negative character count being zero; completely sanitised) as a measure of how ready these solutions are for autonomous deployment.

## LOGICAL: Pipeline Deployment and Feasibility

After benchmarking the performance of the finetuned model, we have developed a Python-based tool that deliberately follows HIL design. This tool has been tested on Windows and macOS and is publicly shared on GitHub (https://github.com/Lekhansh/LOGIC-Local-

Obfuscation-by-GLINER-for-Impartial-Context-Aware-Lineage-). A web-based interface allows PII identification by the model, review by the human and anonymisation with two schemes.

First, the user can choose to maintain the lineage of PIIs within a single note or across a set of notes. These serve separate but related functions. Maintaining relationships within the note keeps the note interpretable, whereas maintaining a dictionary of placeholders across notes allows patterns to emerge in a large corpus of EHR data (Figure 1).

The pipeline includes maintaining corpus-level replacement lists, which inherit their members based on entity type, and fuzzy matching of replaced strings.

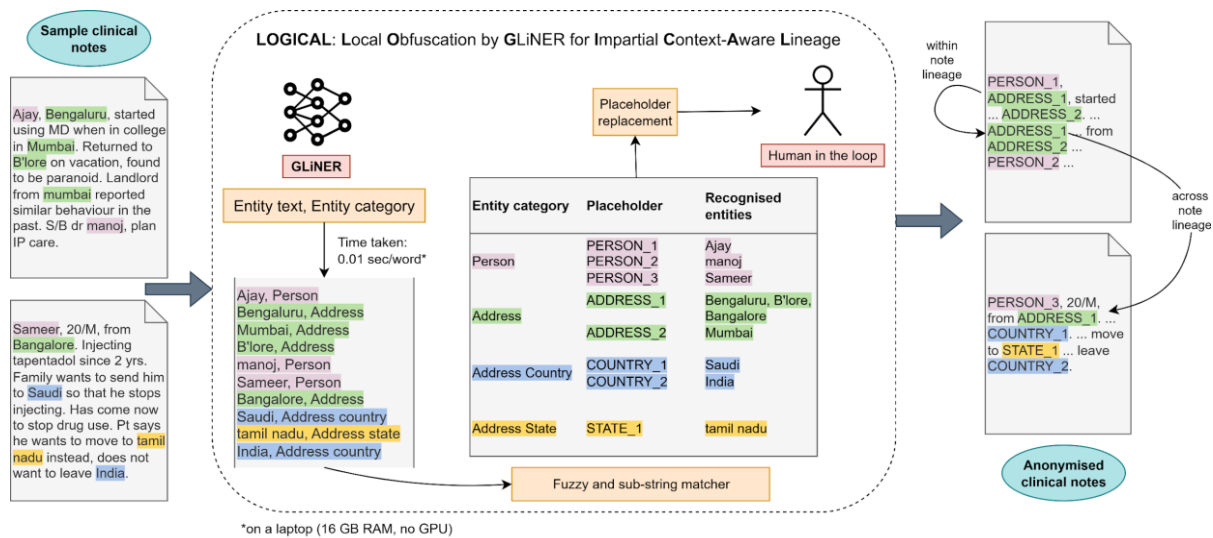

*Figure 1: Schematic of the developed pipeline for the removal of PII while preserving lineage and context*

# Results

More than three-fourths of the instances had PII, with the most common being person and address. We did not find any email or URL addresses, and this category is not considered further. Table 1 shows the prevalence of PII entities in the train and test sets. Supplementary Table 06 has details on the total and unique number of each entity type.

*Table 1: Distribution of PII entities in the dataset and splits*

| Characteristics | Train (N = 2849) | Test (N = 376) |
|---|---|---|
| Instances with no entities, N (%) | 556 (19.5) | 76 (20.2) |
| Prevalence [a], N (%) | | |
| Person | 1788 (62.8) | 249 (62.2) |
| Address | 1462 (51.3) | 148 (39.4) |
| Company | 1395 (49) | 186 (50) |
| Dates | 795 (27.9) | 61 (16.2) |
| Address state | 481 (16.9) | 55 (14.6) |
| Groups | 159 (5.6) | 11 (2.9) |
| Address country | 155 (5.4) | 15 (4) |
| Languages | 88 (3.1) | 8 (2.1) |
| Identification Number | 80 (2.8) | 4 (1.1) |
| Email or URL | 0 (0) | 0 (0) |
| [a] Instances with one or more of a given entity type | | |

Despite a relatively small number of examples, we observe that a finetuned GLiNER model outperforms state-of-the-art LLMs and other solutions. Gemini-Pro-2.5 achieves highly competitive performance in this zero-shot task, performing almost as well as the finetuned GLiNER model on most entity types. Table 2 presents a detailed comparison of the performance of five solutions.

*Table 2: Micro-average measures of character-level performance for PII detection in Electronic Health Records (N = 376)*

| Entity | Metric | GLiNER [a] | Azure [b] | Gemini [c] | Llama [d] | Presidio |
|---|---|---|---|---|---|---|
| **Overall** [e] (detection of PII characters) | Precision | **0.979** | 0.391 | 0.825 | 0.779 | 0.134 |
| | Recall | **0.981** | 0.702 | 0.866 | 0.776 | 0.655 |
| | F1 score | **0.980** | 0.502 | 0.845 | 0.778 | 0.223 |
| | AUROC | **0.990** | 0.839 | 0.931 | 0.886 | 0.781 |
| | Accuracy | **0.999** | 0.971 | 0.994 | 0.991 | 0.905 |
| **Individual entities** | | | | | | |
| Address | Precision | **0.997** | 0.460 | 0.962 | 0.933 | 0.655 |
| | Recall | **1.000** | 0.746 | 0.849 | 0.701 | 0.570 |
| | F1 score | **0.999** | 0.569 | 0.902 | 0.800 | 0.610 |
| | AUROC | **1.000** | 0.872 | 0.920 | 0.850 | 0.785 |
| Address country | Precision | **1.000** | 0.220 | 0.930 | 0.782 | **1.000** |
| | Recall | **1.000** | 0.796 | 0.930 | 0.822 | 0.733 |
| | F1 score | **1.000** | 0.344 | 0.930 | 0.801 | 0.846 |
| | AUROC | **1.000** | 0.898 | 0.960 | 0.911 | 0.866 |
| Address state | Precision | 0.966 | 0.662 | 0.961 | 0.830 | **1.000** |
| | Recall | **1.000** | 0.289 | 0.911 | 0.704 | 0.642 |
| | F1 score | **0.982** | 0.403 | 0.935 | 0.762 | 0.782 |
| | AUROC | **1.000** | 0.645 | 0.960 | 0.852 | 0.821 |
| Company | Precision | **0.944** | 0.688 | 0.698 | 0.895 | 0.138 |
| | Recall | **0.979** | 0.555 | 0.891 | 0.783 | 0.544 |
| | F1 score | **0.961** | 0.615 | 0.783 | 0.835 | 0.220 |
| | AUROC | **0.989** | 0.777 | 0.940 | 0.891 | 0.761 |
| Dates | Precision | **0.987** | 0.067 | 0.774 | 0.299 | 0.022 |
| | Recall | **0.991** | 0.855 | 0.835 | 0.831 | 0.868 |
| | F1 score | **0.989** | 0.125 | 0.803 | 0.440 | 0.043 |
| | AUROC | **0.995** | 0.919 | 0.920 | 0.914 | 0.893 |
| Groups | Precision | **1.000** | NA | 0.157 | 0.500 | 0.170 |
| | Recall | **1.000** | NA | **1.000** | **1.000** | **1.000** |
| | F1 score | **1.000** | NA | 0.271 | 0.667 | 0.290 |
| | AUROC | **1.000** | NA | **1.000** | **1.000** | **1.000** |
| Identification number | Precision | **0.889** | NA | 0.192 | 0.082 | 0.122 |
| | Recall | 0.828 | NA | **1.000** | **1.000** | 0.828 |
| | F1 score | **0.857** | NA | 0.322 | 0.151 | 0.212 |
| | AUROC | 0.914 | NA | **1.000** | **1.000** | 0.914 |
| Languages | Precision | **1.000** | NA | **1.000** | 0.970 | NA |
| | Recall | **0.917** | NA | 0.896 | 0.812 | NA |
| | F1 score | **0.957** | NA | 0.945 | 0.857 | NA |
| | AUROC | **0.958** | NA | 0.948 | 0.906 | NA |
| Person | Precision | **0.999** | 0.846 | 0.973 | 0.939 | 0.319 |

| Entity | Metric | GLiNER [a] | Azure [b] | Gemini [c] | Llama [d] | Presidio |
|--------|--------|-----------|-----------|-----------|-----------|----------|
| | Recall | **0.970** | 0.644 | 0.821 | 0.748 | 0.475 |
| | F1 score | **0.984** | 0.731 | 0.891 | 0.832 | 0.382 |
| | AUROC | **0.985** | 0.822 | 0.910 | 0.874 | 0.733 |

[a] finetuned; [b] Microsoft Azure Text Analytics NER service; [c] Gemini-Pro-2.5; [d] Llama-3.3-70b-Instruct ; [e] adjusted if a solution does not support a particular entity-type

Finetuned GLiNER outperforms at the level of individual instances too (macro-averages) in all categories (Supplementary table 07). However, none of the solutions studied here can be considered ready for a completely autonomous deployment. Finetuned GLiNER could completely sanitise 356 of 376 instances (95%), Gemini-Pro-2.5 (64%) and Llama-3.3-70-b-instruct (54%). Supplementary Table 08 details the performance of all solutions.

On a more cautious note, the finetuned GLiNER model, despite having a very low entity-level FNR of 2%, missed 17 names of persons. Notwithstanding this, finetuned GLiNER had the lowest entity-level FNR in all categories, except for identification numbers, where Gemini-Pro-2.5 performed better (Supplementary Table 09).

We provide two fictitious clinical notes, along with the output of all five solutions compared in this study, in Supplementary Figures 2 and 3.

Finally, the provided LOGICAL pipeline is benchmarked to process clinical text of up to 1000 words in under 15 seconds on a 16 GB RAM laptop without a GPU. The time taken per document depends on its size, with the mean being 0.01 seconds per word and a standard deviation of 0.01 seconds. The maximum time required for a 1000-word document is 100 seconds.

# Discussion

We present the findings of developing and evaluating a PII-masking tool that leverages the performance gains and flexibility provided by transformer-based models without the financial, time, or energy costs associated with LLM-based solutions. Trained on a relatively small dataset of 1515 real-world clinical documents, LOGICAL outperforms state-of-the-art LLMs and proprietary solutions. The overall detection accuracy of LOGICAL (F1 score: 0.98) is significantly higher than that of the next-best solution, Gemini-Pro-2.5 (F1 score: 0.84).

We build on the work by Wiest et al., where the authors      have benchmarked the performance of various open-source LLMs for PII detection (6). LLMs indeed perform remarkably well in zero-shot extraction of PII, with state-of-the-art solutions maintaining their performance even when faced with misspelt, abbreviated, or agrammatical texts. While the specific LLM—Llama-3.3-70b-Instruct—used by Wiest et al. did not perform as well on our dataset, it still outperformed rule-based or hybrid solutions such as the Azure NER service and Microsoft Presidio. Nevertheless, even with aggressive quantisation, deploying a solution that uses a locally hosted LLM with 70 billion parameters requires a GPU with 48 GB VRAM, as suggested by the same group. With a global shortage of GPU machines, such resources are unlikely to be available in most hospital settings (16).

On the other hand, LOGICAL can operate on laptops, making it possible to perform 'sanitisation at source'. It can also enable organisations to be self-reliant when carrying out PII masking at scale for research or compliance purposes. We specifically emphasise the HIL-design because it is a legal and often ethical requirement in anonymisation exercises. For instance, the Safe Harbour method recommended by HIPAA requires a human to determine whether any PII remains after removing 18 PII categories. In HIL scenarios, processing time is crucial, and GLiNER models are faster than LLMs.

In this research, we also aim to expand the scope of PII removal beyond simple redaction, making deidentification a careful process of preserving context, meaning, and signal while maintaining confidentiality. This is an active area of investigation and is particularly relevant to clinical documentation and its applications for training language models. For example, Jiang et al. have proposed Generative Data Refinement (GDR), which uses LLMs to rewrite data to eliminate PII (Jiang et al., 2025).

To train LLMs for clinical tasks, we need large datasets of real-world medical records that include essential clinical signals, such as the timeline of symptoms and the typical age at which illnesses begin, while masking PII. For such applications, we require discreet and relationship-preserving methods of deidentification. LOGICAL represents a step in this direction. It is with this downstream application that we also include PII, which, although not directly leading to identification, could introduce bias, such as religion and language.

This research has several limitations. Most importantly, it may not generalise to other settings. However, we demonstrate that research groups or institutions can choose to invest in annotating a small dataset and train open-source models. The PIIs or other entities can be chosen based on the downstream task and prevailing social, legal and ethical norms. The upfront investment of time in annotation is more than recovered as these finetuned models can achieve almost perfect accuracy on a number of PII types. We reiterate that our aim was not to achieve HIPAA or GDPR compliance, and therefore, this solution does not automate these tasks.

With these limitations in mind, we believe that this research shows the feasibility of using specialised transformer-based models as components of a larger pipeline to sanitise data, enabling pooling of diverse datasets for downstream applications.

# Acknowledgements

We acknowledge the contributions of all doctors and students at the Centre for Addiction Medicine to this work.

# Funding Declaration

Extramural grant from the Indian Council of Medical Research (ICMR), Government of India.

# Supplementary Materials

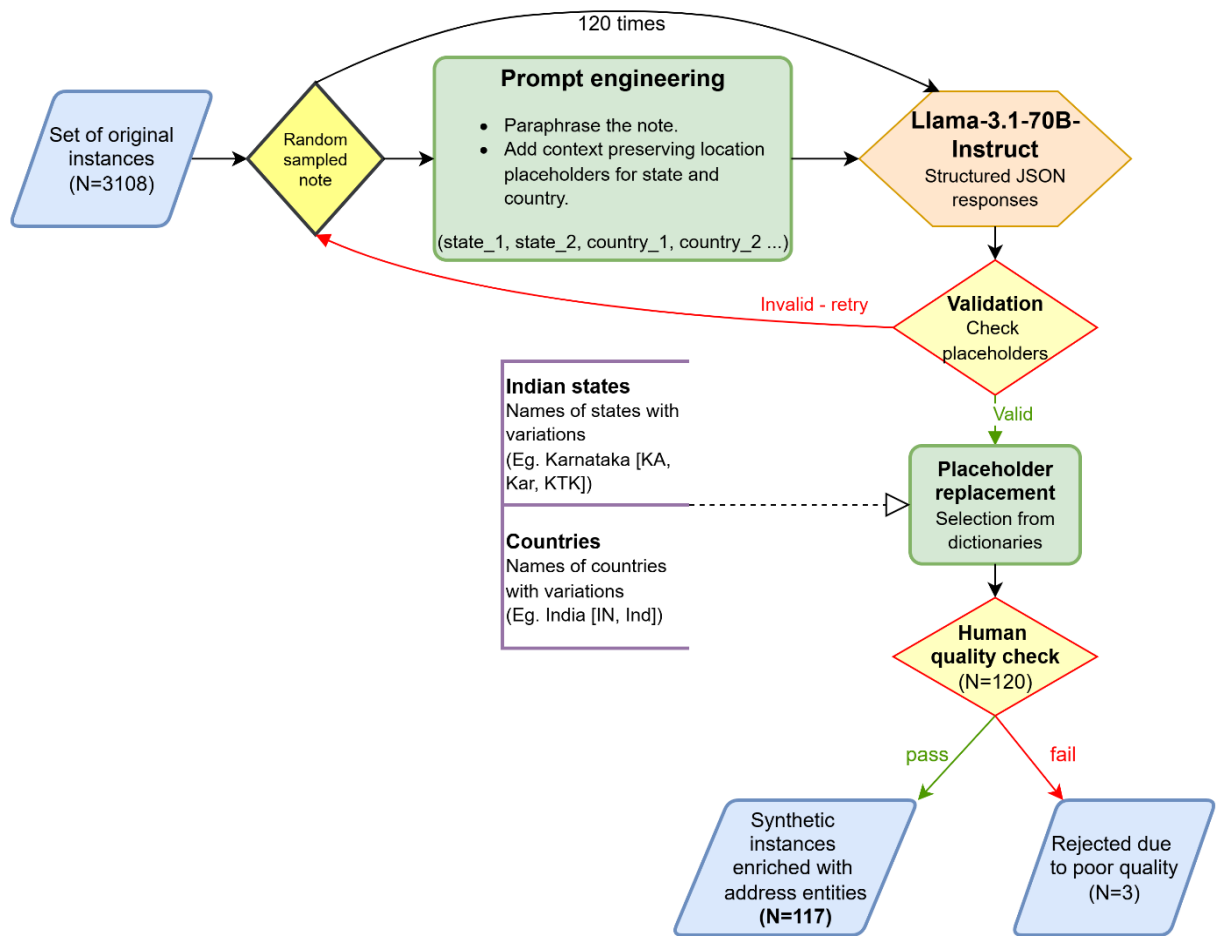

sFigure 1: Synthetic data generation conditioned on real-world data for enriching address-related PII entities

| Hardware | Single GPU NVIDIA RTX A6000 with 48GB VRAM. |
|---|---|
| Software Stack | OS: Ubuntu 22.04 LTS<br>CUDA: 12.2; Python: 3.11.11<br>Libraries: 'gliner' and 'torch' |
| Learning rate main | 0.000003 (This decides the speed of learning in modern BERT layers, and it is deliberately kept very low because modern BERT is already trained on trillions of words, and we do not want it to lose that knowledge – catastrophic forgetting. This knowledge is essential for generalisation to new entities). |
| Learning rate token classifier layers | 0.00003 (This is ten times the above rate. The final layers need more adaptation to this task) |
| Focal loss alpha | 0.75 (This is a static loss scaling factor that gives more importance to less frequent entities. For example, the least frequent entity will get 3 times the importance than the most frequent entity in this setup.) |
| Focal loss gamma | 2 (This parameter scales the loss based on how confident the model is in its predictions. It effectively decreases the importance of entities which the model is already good at detecting and classifying, shifting the focus of training to entities where the model is less confident.) |
| Weight decay | 0.03 for modern BERT layers and 0.01 for token classifier layers. (Weight decay is a regularisation parameter to prevent overfitting on the training data at the cost of generalisability). |
| Learning rate scheduler | Linear (this linearly decreases the learning rate as the training progresses. This allows gentler updates later in the training when the model has already started performing well) |
| Number of epochs | 8 (each epoch refers to passing the whole training data once, with a low learning rate, more epochs are needed. Generally, training is continued till the loss on the test set plateaus) |
| Total runtime | 65 minutes |

sTable 1: Details of finetuning a GLiNER model for PII detection in clinical documents

| Study Entity Labels | Azure Category Label | Azure Sub-Category Label | Azure Label Description |
|---|---|---|---|
| Address | Address | All | A distinct identifier assigned to a physical or geographic location, utilised for navigation, delivery services, and formal administrative purposes. |
| | Location | City | A settlement characterised by a dense population and infrastructure. |
| | Location | Structural | The configuration or organisational schema of components within a system or object that defines the overall architecture. |
| | Location | Geographical | Earth's physical geography and natural features include landforms like rivers, mountains, and valleys. |
| | Location | Location | A specific point or area in physical or virtual space defined by exact coordinates, metadata, or unique identifiers that can be referenced, queried, or accessed. |
| Address state | Location | State | The institutional framework and governing apparatus for a defined geographical area or political entity. |
| Address country | Location | CountryRegion | A distinct territorial entity recognised as a nation or administrative area. |
| | Location | Continent | A vast, continuous landmass on the Earth's surface. |
| | Location | GPE | A geopolitical entity (GPE) is a region or area defined by political boundaries or governance. |
| Dates | DateTime | Date | A specific calendar day expressed in terms of day, month, and year. |
| | DateTime | DateRange | A span of time defined by a start and end date. |
| | DateTime | DateTime | A data type encompassing date and time components used in scheduling or logging events. |
| | DateTime | DateTimeRange | A period defined by a starting and ending date and time. |
| Company | Organization | All | A company, institution, or group formed for a specific purpose. |
| Person | Person | All | An individual human being or a legal entity with rights and responsibilities. |
| Identification number | | No matching categories | |
| Languages | | No matching categories | |
| Groups | | No matching categories | |

sTable 2: Mapping of PII categories used in the study and Microsoft Azure Text Analytics NER service

| Study entity labels | Microsoft Presidio Entity Types | Presidio Label Description |
|---|---|---|
| Person | PERSON | A full person name, which can include first names, middle names or initials, and last names. |
| Address, Address state, Address country | LOCATION | Name of a politically or geographically defined location, such as cities, provinces, countries, international regions, bodies of water, or mountains. (If correctly labelled as a location, we have counted it as accurate without regard to finer categorisations). |
| Dates | DATE_TIME | Absolute or relative dates, periods, or times smaller than a day. |
| Groups | NRP | A person's nationality, religious or political group. |
| Company | ORGANIZATION | Companies, agencies, institutions, and other organisations. |
| Identification number | US_DRIVER_LICENSE | A US driver's license, according to https://ntsi.com/drivers-license-format/ |
| | US_PASSPORT | A US passport number with nine digits. |
| | CREDIT_CARD_NUMBER | A credit card number is between 12 and 19 digits. |
| | IBAN_CODE | The International Bank Account Number (IBAN) is an internationally agreed system of identifying bank accounts across national borders to facilitate the communication and processing of cross-border transactions with a reduced risk of transcription errors. |
| | PHONE_NUMBER | A telephone number |
| | MEDICAL_LICENSE | Standard medical license numbers. |
| | IN_PAN | The Indian Permanent Account Number (PAN) is a unique 12-character alphanumeric identifier issued to all business and individual entities registered as taxpayers. |
| | IN_AADHAAR | The Indian government issued a unique 12-digit individual identity number. |
| | IN_VOTER | The Indian Election Commission issued 10 10-digit alphanumeric voter ID for all Indian citizens (age 18 or above) |
| | IN_VEHICLE_REGISTRATION | The Indian government issued transport (govt, personal, diplomatic, defence) vehicle registration numbers |
| | IN_PASSPORT | Indian Passport Number |
| Languages | No matching categories | Not applicable |

sTable 3: Mapping of PII categories used in the study and Microsoft Presidio

## Prompt for Extraction of PII

You are an API that helps researchers in sanitising clinical notes. Given between three backticks is a clinical note, process it as per the <Instructions> and return your response as JSON mentioned in <Response_Schema>.

<Instructions>
You need to extract some entities from this text. The types of entities are mentioned in the <Entities> section along with their description and examples.
1. Search the note for each entity type one by one.
2. If you find a particular type extract the word or phrase. If the words are contiguous then return it as a single phrase but if they are not contiguous then return them separately.
3. If an entity is not found, return nothing.
4. Ensure you make no changes to the words or phrases you found.
5. Preserve the case, any punctuations etc that are present in the extracted phrases.
6. Your work will be checked programmatically by searching for an exact match in the text. So do not alter the extracted entity at all otherwise you will fail. For example, "bangalore" should not become "Bangalore".
7. These notes are not proofread and will have all types of abbreviations, misspellings etc. You must infer the presence and type of entity based on the context.
8. You must infer the presence and type of entity based on context as many of them may be new to you or misspelt. See "Specific Instructions" and "Examples" for each entity type.
</Instructions>
<Entities>
<person>
# Description:
These are names of persons.
# Specific Instructions
1.Extract all parts of the name.
# Examples
Dr LS, Anil, Mini, Teena SNM.
</person>
<address country>
# Description
These are names of countries. They may be abbreviated.
</address country>
<address state>
# Description
These are names of Indian states or equivalent larger geographical areas in other countries.
# Specific Instructions:
1.Indian state names are frequently abbreviated, and you must the context to infer their presence.
2.Some states like "Tamil Nadu" consist of two words, you must extract the full name.
</address state>
<address>
# Description
These are names of locations and geographical entities that are smaller than states or equivalent categories in other countries.
# Specific Instructions:
1.Indian city names are frequently abbreviated, and you must use the context to infer their presence.
2.Location names may be new to you, misspelt or abbreviated. You must rely on the context.
# Examples
{Jayanagar, BTM Layout, Dasarahalli, Uttarahalli, Benagluru}
</address>
<company>
# Description
These are names of Health care facilities (HCF), places or work and employers.
# Specific Instructions
1.Extract only if there is a name to the entity which can lead to identification, for example "government hospital" is a generic term and need not be included.
2.These names can be abbreviated or have many words so extract carefully and check your work.
# Examples
{Centre for Addiction Medicine (CAM),BHEL,Indian Railways,St. John's,AIIMS,Kasturba Medical College,NIMHANS, CMC}
</company>
<languages>
# Description
These are names of languages that may be abbreviated but can be inferred from the context.
</languages>
<groups>
# Description
These are names of religions, castes, political or other types of groups which can reveal a person's identity.
# Examples
{hindu, hakki-pikki, BJP}
</groups>
<dates>
# Description
These are fully specified dates i.e. where day of the month, month and year are all present or inferred from the context.
# Specific Instructions
1.Extract only if the full date is specified or can be inferred. For example "26/11/2023" or "23rd to 26 November 2023".
</dates>
<identification number>
# Description
These are numeric or alphanumeric identifiers. For example, phone numbers, Driving license, PAN Card or Aadhaar Card numbers. This also includes identifiers of hospital records like Patient ID or Unique IDs.

**Prompt for Extraction of PII**

```
</identification number>
</Entities>
<Response_Schema>
{
"person": [],
"address country":[],
"address state":[],
"address":[].
"company":[],
"languages"[],
"groups":[],
"dates":[],
"identification number":[]
}
</Response_Schema>
```replaceMeWithInputText```
```

sTable 4: Prompt used for the extraction of PII with Gemini-Pro-2.5 and Llama-3.1-70b-Instruct

| PII Type | Heuristics for pseudonymisation |
|---|---|
| Person | ➢ Keep the honorific/title like "Dr/Mr/Ms." as is.<br>➢ Replace with the same number of words.<br>➢ If multiple capitalised initials like "SNM" or "GVL", use the random string generator to replace them with a random string of the same length, checking that the replacement is not the same as the replaced.<br>➢ For non-abbreviated First and Second names, use the random Indian name generator on the dashboard.<br>➢ Do not try to match gender.<br>➢ Do not change case. If the name was not adequately capitalised, keep the replacement also non-capitalised.<br>➢ Ensure that all parts of the name, except the title, have been replaced. |
| Address State | ➢ If abbreviated, use the random Indian state abbreviation generator to replace with the same character length.<br>➢ If misspelt, replace with a misspelt version of an Indian state.<br>➢ If multi-word, the number of words in the replacement must be the same as the number of words in the original.<br>➢ Do not change case, maintain capitalisation errors if any.<br>➢ Ensure all parts of the state name have been replaced. |
| Address Country | ➢ All rules mentioned above apply, with the following exceptions:<br>➢ Use a random country abbreviation generator. |
| Address | ➢ If a city name, replace it with another city name with the same number of words.<br>➢ If a locality name is replaced with another locality name with the same number of words.<br>➢ Ensure all parts of the name have been replaced. |
| Identification Number | ➢ If alphanumeric, use the alphanumeric random generator specifying the number of alphabets, the number of numeric characters, and the case.<br>➢ If only numeric, use the random numeric generator to replace with same number of characters.<br>➢ If hyphenated, maintain hyphenation by placing the hyphen in the right position. |
| Company | ➢ If the name of a health facility is given, then replace it with the name of another health facility that has the most similar composition in terms of the number of words, presence of abbreviations or contractions, and occurrence of the word "hospital".<br>➢ If there were spelling mistakes, let the replacement have a comparable number of words misspelt.<br>➢ If a place of occupation, replace from the list which has the most similar composition in terms of the number of words and presence of abbreviations. |
| Dates | ➢ Use the random date generator and replace it in the same format as the original.<br>➢ If it is a date range, replace it with a range using two random generations. |
| Languages | Do not replace. |
| Groups | |

sTable 5: Heuristics and rules for pseudonymising the test instances before comparison of PII removal solutions

| Entity Type | Count [a] | All Word Count [b] | All Character Count [b] |
|---|---|---|---|
| person | 4372 (2531) | 2.0 (1.0, 2.0); 1.85 | 10.0 (8.0, 13.0); 11.22 |
| address | 2723 (1189) | 1.0 (1.0, 2.0); 1.54 | 9.0 (8.0, 11.0); 11.42 |
| company | 2653 (725) | 2.0 (1.0, 3.0); 2.38 | 9.0 (3.0, 24.0); 15.47 |
| dates | 1424 (1065) | 1.0 (1.0, 1.0); 1.49 | 10.0 (8.0, 10.0); 9.44 |
| address state | 688 (202) | 1.0 (1.0, 2.0); 1.33 | 9.0 (6.0, 10.0); 8.45 |
| address country | 317 (114) | 1.0 (1.0, 1.0); 1.21 | 5.0 (5.0, 8.0); 6.35 |
| groups | 181 (51) | 1.0 (1.0, 1.0); 1.15 | 5.0 (5.0, 9.0); 7.13 |
| identification number | 126 (122) | 2.0 (1.0, 3.0); 2.12 | 10.0 (8.0, 15.0); 11.62 |
| languages | 116 (29) | 1.0 (1.0, 1.0); 1.03 | 7.0 (5.0, 7.0); 6.57 |

[a] Total (Unique); [b] Median (Q1, Q3); Mean.

sTable 6: Details of PII-entities in the whole dataset

| Entities | Metric[a] | GLiNER (finetuned) | Microsoft Azure Text Analytics NER service | Gemini-Pro-2.5 | Llama-3.1-70b-Instruct | Microsoft Presidio |
|---|---|---|---|---|---|---|
| **Overall** | Precision | **1** (0.92, 1) | 0.36 (0.24, 0.55) | 0.9 (0.48, 1) | 0.89 (0.15, 1) | 0.12 (0.06, 0.21) |
| | Recall | **1** (0.89, 1) | 0.58 (0.44, 0.73) | 0.91 (0.32, 1) | 0.75 (0.15, 1) | 0.68 (0.43, 0.86) |
| | F1 score | **1** (0.88, 1) | 0.41 (0.31, 0.56) | 0.84 (0.42, 1) | 0.75 (0.19, 0.97) | 0.2 (0.11, 0.32) |
| | AUROC | **1** (0.94, 1) | 0.77 (0.7, 0.85) | 0.95 (0.66, 0.99) | 0.87 (0.57, 1) | 0.8 (0.68, 0.88) |
| **Individual entities** | | | | | | |
| Person | Precision | **1** (1, 1) | **1** (0.75, 1) | **1** (1, 1) | **1** (0.93, 1) | 0.26 (0, 0.6) |
| | Recall | **1** (1, 1) | 0.62 (0.36, 0.77) | **1** (0.76, 1) | **1** (0.5, 1) | 0.41 (0, 0.67) |
| | F1 score | **1** (1, 1) | 0.7 (0.51, 0.85) | **1** (0.83, 1) | 0.96 (0.66, 1) | 0.33 (0, 0.58) |
| Address | Precision | **1** (1, 1) | 0.53 (0, 1) | **1** (1, 1) | **1** (0.75, 1) | 0.45 (0, 1) |
| | Recall | **1** (1, 1) | **1** (0, 1) | **1** (1, 1) | **1** (0.46, 1) | 0.41 (0, 1) |
| | F1 score | **1** (1, 1) | 0.67 (0, 1) | **1** (1, 1) | **1** (0.53, 1) | 0.5 (0, 1) |
| Address state | Precision | **1** (1, 1) | 0 (0, 0.25) | **1** (1, 1) | **1** (0, 1) | **1** (0, 1) |
| | Recall | **1** (1, 1) | 0 (0, 0.1) | **1** (1, 1) | **1** (0, 1) | **1** (0, 1) |
| | F1 score | **1** (1, 1) | 0 (0, 0.14) | **1** (1, 1) | **1** (0, 1) | **1** (0, 1) |
| Address country | Precision | **1** (1, 1) | 0 (0, 0) | **1** (1, 1) | **1** (0.9, 1) | **1** (1, 1) |
| | Recall | **1** (1, 1) | 0 (0, 0) | **1** (0.89, 1) | **1** (0.77, 1) | **1** (0.77, 1) |
| | F1 score | **1** (1, 1) | 0 (0, 0) | **1** (0.94, 1) | **1** (0.82, 1) | **1** (0.87, 1) |
| Dates | Precision | **1** (1, 1) | 0 (0, 0) | **1** (0.85, 1) | 0 (0, 1) | 0 (0, 0) |
| | Recall | **1** (1, 1) | 0 (0, 0) | **1** (0.32, 1) | 0 (0, 1) | 0 (0, 0) |
| | F1 score | **1** (1, 1) | 0 (0, 0) | **1** (0.48, 1) | 0 (0, 1) | 0 (0, 0) |
| Company | Precision | **1** (1, 1) | 0 (0, 0) | **1** (0.37, 1) | **1** (0, 1) | 0 (0, 0.7) |
| | Recall | **1** (1, 1) | 0 (0, 0) | **1** (0.59, 1) | 0.84 (0, 1) | 0 (0, 0.36) |
| | F1 score | **1** (1, 1) | 0 (0, 0) | 0.91 (0.5, 1) | 0.82 (0, 1) | 0 (0, 0.11) |
| Identification number | Precision | **1** (0, 1) | Not supported | 0 (0, 0.31) | 0 (0, 0.47) | 0 (0, 0) |
| | Recall | **1** (0, 1) | | 0 (0, 0.5) | 0 (0, 0.75) | 0 (0, 0) |
| | F1 score | **1** (0, 1) | | 0 (0, 0.38) | 0 (0, 0.58) | 0 (0, 0) |
| Languages | Precision | **1** (1, 1) | Not supported | **1** (1, 1) | **1** (1, 1) | Not supported |
| | Recall | **1** (1, 1) | | **1** (1, 1) | **1** (0, 1) | |
| | F1 score | **1** (1, 1) | | **1** (1, 1) | **1** (0, 1) | |
| Groups | Precision | **1** (1, 1) | Not supported | 0 (0, 1) | **1** (0, 1) | 0 (0, 0) |
| | Recall | **1** (1, 1) | | 0 (0, 1) | **1** (0, 1) | 0 (0, 0) |
| | F1 score | **1** (1, 1) | | 0 (0, 1) | **1** (0, 1) | 0 (0, 0) |

sTable 7: Comparison of finetuned GLiNER for PII detection with other solutions using macroaverage metrics

| PII-detection Solution | Instances processed | Completely sanitised [a] | Sanitisation percentage | Total entities [b] | Completely missed entities | Entity miss-rate percentage |
|---|---|---|---|---|---|---|
| GLiNER (finetuned) | 376 | 356 | **94.68** | 1151 | 25 | **2.17** |
| Microsoft Azure Text Analytics NER service | 300 | 27 | 9 | 1127 | 406 | 36.02 |
| Gemini-Pro-2.5 | 376 | 241 | 64.1 | 1151 | 125 | 10.86 |
| Llama-3.1-70b-Instruct | 376 | 203 | 53.99 | 1151 | 288 | 25.02 |
| Microsoft Presidio | 300 | 40 | 13.33 | 1142 | 488 | 42.73 |

[a] defined as no characters belonging to an entity being missed; [b] adjusted for lack of comparable entities in Azure and Presidio

sTable 8: Performance of PII-detection solutions for completely removing all PII categories of interest in a single pass

| Entity | GLiNER(finetuned) | | Microsoft Azure Text Analytics NER service | | Gemini-Pro-2.5 | | Llama-3.1-70b-Instruct | | Microsoft Presidio | |
|---|---|---|---|---|---|---|---|---|---|---|
| | Total (missed) | % | Total (missed) | % | Total (missed) | % | Total (missed) | % | Total (missed) | % |
| Address | 200 (0) | **0** | 200 (48) | 24 | 200 (25) | 12.5 | 200 (59) | 29.5 | 200 (80) | 40 |
| Address-country | 26 (0) | **0** | 26 (5) | 19.23 | 26 (0) | **0** | 26 (3) | 11.54 | 26 (6) | 23.08 |
| Address-state | 70 (0) | **0** | 70 (43) | 61.43 | 70 (4) | 5.71 | 70 (21) | 30 | 70 (24) | 34.29 |
| Company | 307 (5) | **1.63** | 307 (215) | 70.03 | 307 (30) | 9.77 | 307 (94) | 30.62 | 307 (192) | 62.54 |
| Dates | 89 (1) | **1.12** | 89 (14) | 15.73 | 89 (11) | 12.36 | 89 (11) | 12.36 | 89 (7) | 7.87 |
| Groups | 11 (0) | **0** | Not available | | 11 (0) | **0** | 11 (0) | **0** | 11 (0) | **0** |
| Identification number | 4 (1) | 25 | Not available | | 4 (0) | **0** | 4 (0) | **0** | 4 (1) | 25 |
| Languages | 9 (1) | **11.11** | Not available | | 9 (1) | **11.11** | 9 (2) | 22.22 | Not available | |
| Person | 435 (17) | **3.91** | 435 (81) | 18.62 | 435 (54) | 12.41 | 435 (98) | 22.53 | 435 (178) | 40.92 |

sTable 9: Performance of PII-detection solutions - completely missing an entity

**LOGICAL**
Rajiv, 22, m, from Sandesh nagar, Blore, studied till 10th, works as delivery agent. has come with mother Mrs Rathnamma. tapentadol IV use since 2 years, started with college friends. injects 800-900mg per day. developed craving, tolerance, loss of control and withdrawal within 1 month of use, has been suspended from college due to drug use, but contd to use despite failing semester. has also developed ulcers and needed skin grafting. demanding money from family to procure tabs, physical aggression on some occasions. was then sent to Hyderabad to stay with relatives, but returned to bangalore within a month. has been brought here to BTM hospital for treatment. injection marks - both forearms. COWS=4.
imp: ODS withdrawal (IDU- tapentadol).
plan: doesnt want to get admitted today, will come on 03-9-22. requesting for OST on OPD basis, informed about rules, consent form signed. ID details taken - aadhar num: 111111111111. OST being started with bup 0.4mg SL, f/u tomorrow.
case discussed with Dr Krishna. imp and plan concurred.

**Llama-3.3-70b-Instruct**
Rajiv, 22, m, from Sandesh nagar, Blore, studied till 10th, works as delivery agent. has come with mother Mrs Rathnamma. tapentadol IV use since 2 years, started with college friends. injects 800-900mg per day. developed craving, tolerance, loss of control and withdrawal within 1 month of use, has been suspended from college due to drug use, but contd to use despite failing semester. has also developed ulcers and needed skin grafting. demanding money from family to procure tabs, physical aggression on some occasions. was then sent to Hyderabad to stay with relatives, but returned to bangalore within a month. has been brought here to BTM hospital for treatment. injection marks - both forearms. COWS=4.
imp: ODS withdrawal (IDU- tapentadol).
plan: doesnt want to get admitted today, will come on 03-9-22. requesting for OST on OPD basis, informed about rules, consent form signed. ID details taken - aadhar num: 111111111111. OST being started with bup 0.4mg SL, f/u tomorrow.
case discussed with Dr Krishna. imp and plan concurred.

**Gemini 2.5 Pro**
Rajiv, 22, m, from Sandesh nagar, Blore, studied till 10th, works as delivery agent. has come with mother Mrs Rathnamma. tapentadol IV use since 2 years, started with college friends. injects 800-900mg per day. developed craving, tolerance, loss of control and withdrawal within 1 month of use, has been suspended from college due to drug use, but contd to use despite failing semester. has also developed ulcers and needed skin grafting. demanding money from family to procure tabs, physical aggression on some occasions. was then sent to Hyderabad to stay with relatives, but returned to bangalore within a month. has been brought here to BTM hospital for treatment. injection marks - both forearms. COWS=4.
imp: ODS withdrawal (IDU- tapentadol).
plan: doesnt want to get admitted today, will come on 03-9-22. requesting for OST on OPD basis, informed about rules, consent form signed. ID details taken - aadhar num: 111111111111. OST being started with bup 0.4mg SL, f/u tomorrow.
case discussed with Dr Krishna. imp and plan concurred.

**Microsoft Presidio**
Rajiv, 22, m, from Sandesh nagar, Blore, studied till 10th, works as delivery agent. has come with mother Mrs Rathnamma. tapentadol IV use since 2 years, started with college friends. injects 800-900mg per day. developed craving, tolerance, loss of control and withdrawal within 1 month of use, has been suspended from college due to drug use, but contd to use despite failing semester. has also developed ulcers and needed skin grafting. demanding money from family to procure tabs, physical aggression on some occasions. was then sent to Hyderabad to stay with relatives, but returned to bangalore within a month. has been brought here to BTM hospital for treatment. injection marks - both forearms. COWS=4.
imp: ODS withdrawal (IDU- tapentadol).
plan: doesnt want to get admitted today, will come on 03-9-22. requesting for OST on OPD basis, informed about rules, consent form signed. ID details taken - aadhar num: 111111111111. OST being started with bup 0.4mg SL, f/u tomorrow.
case discussed with Dr Krishna. imp and plan concurred.

**Microsoft Azure Text Analytics NER service**
Rajiv, 22, m, from Sandesh nagar, Blore, studied till 10th, works as delivery agent. has come with mother Mrs Rathnamma. tapentadol IV use since 2 years, started with college friends. injects 800-900mg per day. developed craving, tolerance, loss of control and withdrawal within 1 month of use, has been suspended from college due to drug use, but contd to use despite failing semester. has also developed ulcers and needed skin grafting. demanding money from family to procure tabs, physical aggression on some occasions. was then sent to Hyderabad to stay with relatives, but returned to bangalore within a month. has been brought here to BTM hospital for treatment. injection marks - both forearms. COWS=4.
imp: ODS withdrawal (IDU- tapentadol).
plan: doesnt want to get admitted today, will come on 03-9-22. requesting for OST on OPD basis, informed about rules, consent form signed. ID details taken - aadhar num: 111111111111. OST being started with bup 0.4mg SL, f/u tomorrow.
case discussed with Dr Krishna. imp and plan concurred.

**Legend**
person
address
address country
dates
company
identification number

sFigure 2: Examples of five PII detection solutions on a fictitious clinical note. Note the missed date and false positive with Microsoft Presidio and Azure Text Analytics.

**LOGICAL**
s/b Chandan, PSW trainee.
Charles, 58/M, first visit post discharge. was admitted in lifeline rehab centre in belgaum for ADS. last use of alcohol 1 month ago.
some meds given in rehab, but details NA. had DT a month ago, managed in rehab with some meds. currently staying in maharashtra
with wife, has come from kolhapur now for f/u visit. restarted working as tailor, wants to resume meds.
seen by Dr Ranjana. restart baclofen upto 60mg i/v/o good resp in the past. f/u after 1 month.

**Llama-3.3-70b-Instruct**
s/b Chandan, PSW trainee.
Charles, 58/M, first visit post discharge. was admitted in lifeline rehab centre in belgaum for ADS. last use of alcohol 1 month ago.
some meds given in rehab, but details NA. had DT a month ago, managed in rehab with some meds. currently staying in maharashtra
with wife, has come from kolhapur now for f/u visit. restarted working as tailor, wants to resume meds.
seen by Dr Ranjana. restart baclofen upto 60mg i/v/o good resp in the past. f/u after 1 month.

**Gemini 2.5 Pro**
s/b Chandan, PSW trainee.
Charles, 58/M, first visit post discharge. was admitted in lifeline rehab centre in belgaum for ADS. last use of alcohol 1 month ago.
some meds given in rehab, but details NA. had DT a month ago, managed in rehab with some meds. currently staying in maharashtra
with wife, has come from kolhapur now for f/u visit. restarted working as tailor, wants to resume meds.
seen by Dr Ranjana. restart baclofen upto 60mg i/v/o good resp in the past. f/u after 1 month.

**Microsoft Presidio**
s/b Chandan, PSW trainee.
Charles, 58/M, first visit post discharge. was admitted in lifeline rehab centre in belgaum for ADS. last use of alcohol 1 month ago.
some meds given in rehab, but details NA. had DT a month ago, managed in rehab with some meds. currently staying in maharashtra
with wife, has come from kolhapur now for f/u visit. restarted working as tailor, wants to resume meds.
seen by Dr Ranjana. restart baclofen upto 60mg i/v/o good resp in the past. f/u after 1 month.

**Microsoft Azure Text Analytics NER service**
s/b Chandan, PSW trainee.
Charles, 58/M, first visit post discharge. was admitted in lifeline rehab centre in belgaum for ADS. last use of alcohol 1 month ago.
some meds given in rehab, but details NA. had DT a month ago, managed in rehab with some meds. currently staying in maharashtra
with wife, has come from kolhapur now for f/u visit. restarted working as tailor, wants to resume meds.
seen by Dr Ranjana. restart baclofen upto 60mg i/v/o good resp in the past. f/u after 1 month.

**Legend**
person
address
address state
dates
company

sFigure 3: Examples of five PII detection solutions on a fictitious clinical note. Note the missed name in Presidio and the high false-positives in the output of Presidio and Azure Text Analytics.